\documentclass[10pt,twocolumn,letterpaper]{article}

\usepackage{wacv}
\usepackage{times}
\usepackage{epsfig}
\usepackage{graphicx}
\usepackage{amsmath}
\usepackage{amssymb}
\usepackage{booktabs}
\usepackage{color}
\usepackage{colortbl}
\definecolor{blue}{rgb}{0,0,1}
\definecolor{ky}{rgb}{0.4,0.7,0}
\newcommand{\ji}[1]{\textcolor{black}{{#1}}}
\newcommand{\ky}[1]{\textcolor{black}{{#1}}}

\usepackage{algorithm2e}
\usepackage{algpseudocode}
\usepackage{float}
\usepackage[dvipsnames]{xcolor}
\usepackage{subcaption}
\usepackage{multirow}

\def\wacvPaperID{1504} 

\wacvfinalcopy 

\ifwacvfinal
\usepackage[breaklinks=true,bookmarks=false]{hyperref}
\else
\usepackage[pagebackref=true,breaklinks=true,colorlinks,bookmarks=false]{hyperref}
\fi

\begin{document}
	
	\title{SimOn: A Simple Framework for Online Temporal Action Localization}
	
\author{Tuan N. Tang, Jungin Park, Kwonyoung Kim, Kwanghoon Sohn \\
	School of Electrical and Electronic Engineering \\
	Yonsel University \\
	{\tt\small \{tuantng, newrun, kyk12, khsohn\}@yonsei.ac.kr
}
}

\maketitle

\begin{abstract}
     \ji{Online Temporal Action Localization (On-TAL) aims to immediately provide action instances from untrimmed streaming videos.
     The model is not allowed to utilize future frames and any processing techniques to modify past predictions, making On-TAL much more challenging.
     In this paper, we propose a simple yet effective framework, termed SimOn, that learns to predict action instances using the popular Transformer architecture in an end-to-end manner.
     Specifically, the model takes the current frame feature as a query and a set of past context information as keys and values of the Transformer.
     Different from the prior work that uses a set of outputs of the model as past contexts, we leverage the past visual context and the learnable context embedding for the current query.
    Experimental results on the THUMOS14 and ActivityNet1.3 datasets show that our model remarkably outperforms the previous methods, achieving a new state-of-the-art On-TAL performance.
    In addition, the evaluation for Online Detection of Action Start (ODAS) demonstrates the effectiveness and robustness of our method in the online setting.
    The code is available at \url{https://github.com/TuanTNG/SimOn}.}
\end{abstract}

\section{Introduction}
    \ji{Video understanding has attracted attention from the computer vision community with the development of video platforms such as YouTube.
    One of the most well-liked tasks is Temporal Action Localization (TAL)~\cite{shou2016temporal, long2019gaussian, yuan2017temporal, lin2017single, buch2019end, xu2020g, zeng2019graph, lin2018bsn, zhao2017temporal, chao2018rethinking, buch2017sst, shou2017cdc} which aims to predict action instances from an untrimmed video.}

    \ji{Most existing TAL models have been trained in an offline fashion.
    In other words, the models are allowed to access the whole frames in a video so that they can take the relationships between all frames into account and apply the post-processing techniques such as Non-maximum suppression (NMS).
    However, they are inherently impractical for real-world applications such as live sports broadcasting where the frames are sequentially provided and the future frames are not accessible.}

    \begin{figure}[t]
    \begin{center}
       \includegraphics[width=1\linewidth]{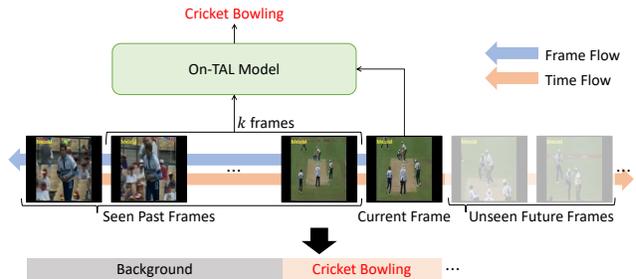}
    \end{center}
       \caption{In On-TAL, video frames are sequentially given to the model to produce action instances.}
    \label{fig:1}
    \end{figure}

    \ji{Recently, online temporal action localization (On-TAL) has been introduced, incorporating TAL into streaming videos~\cite{kang2021cag}.
    In the online setting, the model is \textit{not allowed} to access future frames and can take the past and current frames only.
    In addition, refinement techniques (\eg NMS) also can not be applied to the output of the model, making On-TAL more challenging.
    The prior work~\cite{kang2021cag} has augmented context information into the actionness grouping process by formulating On-TAL as a Markov Decision Process (MDP).
    They have successfully incorporated On-TAL into Q-imitation learning and provided a good baseline for On-TAL.
    However, they trained the whole model in two-staged so that the OAD models that produce the context states and the decision model that determines whether to classify the current frame should be trained separately, making the training procedure inefficient.
    Furthermore, the decision has been made heavily depending on context information without taking visual information into account.}

    \ji{In this paper, we propose a simple framework for On-TAL, termed SimOn, that formulates On-TAL as a sequential prediction problem and learns to immediately predict action instances from a few past information using Transformers~\cite{vaswani2017attention} in an end-to-end manner.
    Unlike the prior work~\cite{kang2021cag} where the outputs of the OAD models are sequentially given as the contexts for decision-making, we leverage past visual information as a long-term context and introduce a learnable context embedding as a short-term context to take the relations of consecutive two frames into account.
    Since we use only a few past frame information, our method can be carried out with high computational and memory efficiency.}
    
    \ji{Extensive experiments prove the superiority and effectiveness of the proposed method, showing state-of-the-art performance for On-TAL on the THUMOS14~\cite{jiang2014thumos} and ActivityNet1.3~\cite{caba2015activitynet} datasets with real-time inference speed.
    In addition, the experiments for ODAS on the THUMOS14 dataset demonstrate the robustness of our method to the online video tasks, outperforming state-of-the-art approaches.}

Our contributions are summarized as follows:
\begin{itemize}
	\item We introduce a simple yet effective framework with a low computational cost for On-TAL. This is the first attempt to learn the On-TAL model in an end-to-end manner that enables efficient training.
	\item We formulate past context information as long-term and short-term contexts to enable the model to predict action instances in challeging conditions.
	\item The proposed method outperforms state-of-the-art approaches for On-TAL and ODAS with large margins, demonstrating the robustness of our method to the online video tasks.
\end{itemize}

\section{Related Work}
\subsection{Temporal Action Localization}
\label{related_work:tal}


The goal of Temporal Action Localization (TAL) is to accurately produce action instances in a video sequence, including the start time, end time, and type of action, in a video sequence. The task has recently been received tremendous attention for from researchers \cite{bai2020boundary, lin2020fast,long2019gaussian, yuan2017temporal, lin2017single, buch2019end, xu2020g, zeng2019graph, lin2018bsn, zhao2017temporal, chao2018rethinking, buch2017sst, shou2017cdc, shou2016temporal, xiong2017pursuit, gao2017turn, yeung2016end, alwassel2018action}. Detailed approaches related to TAL can be found in the survey article \cite{kang2021cag}. One problem of TAL which prevents the model from the real-time application is that the model is allowed to exploit the future frames which is not suitable for real application.

Our task is different and inspired by recent work, CAG-QIL \cite{kang2021cag}, which proposes to generate action in an online fashion. Hence, the model can run in a real-time manner. A more review of current approaches is in section \ref{related_work:on-tal}.

\subsection{Online Video Understanding}
\label{related_work:ovd}
Online Video Understanding can be classified as Online Action Detection (OAD) and Online Detection of Action Start (ODAS). 

The goal of OAD is to determine whether or not an action occurs in the current frame of untrimmed video and to classify the type of action. The output of OAD is the per-frame class probability which is dissimilar to the action instance that contains start time, end time, and category. The work \cite{geest2016online} is the first approach in ODA that classifies the per-clip feature. Following work improve ODA by two-stream network \cite{de2018modeling}, deep LSTM \cite{li2016online}, Temporal Recurrent Network \cite{xu2019temporal}, and Reinforcement Learning \cite{gao2017red}.

Different from OAD, ODAS \cite{shou2018online} aims to immediately detect the start of the action instance. It provides an intermittent start time and is thus more realistic. Many methods try to improve ODAS including \cite{wang2018back} which uses bidirectional LSTM and \cite{gao2019startnet} which applies reinforcement learning and minimizes the reward for start detection prediction.

    While OAD and ODAS share the same spirit with online video tasks, On-TAL has different aspects from some perspectives:
    While OAD and ODAS only provide the per-frame action category and the start time of action instance respectively, On-TAL proposes completed action instances including start time, end time, and type of action as in TAL.

\subsection{On-TAL}
\label{related_work:on-tal}
Recent work CAG-QIL \cite{kang2021cag} introduces a new task, On-TAL, which provides action instances as soon as the action ended. To clarify, the three kinds of tasks including OAD, ODAS, and On-TAL have the same setting, in which the model is not allowed to utilize future frames, but must guarantee the model performance. On-TAL, on the other hand, is much more complicated than OAD and ODAS due to the designed outputs that contain TAL's action instances and is not permitted to use any post-processing techniques. 
CAG-QIL solves On-TAL by first training the OAD model, then post-processing it to get more accurate per-frame action outputs, and finally performing Context-Aware Actionness Grouping to form action instances. CAG-QIL first suggests that OAD outputs be post-processed using standard supervised learning. Overfitting, on the other hand, significantly degrades performance. As a result, CAG-QIL formulates the problem as a Markov Decision Process and employs Deep Q Imitation Learning to correct the output of OAD. CAG-QIL, in addition to contemporary OAD outputs, considers the decision context, which has been shown to be critical for current-frame decisions.

\section{Problem Statement and Motivation}
\subsection{Online Temporal Action Localization}

    \ji{Assume an untrimmed video $V= \lbrace{x_1, x_2, ..., x_T}\rbrace$ with $T$ frames, where $x_t$ is the $t$-th frame. $V$ contains $M$ actions such that the action instances are denoted by $\psi = \{(s_m, e_m, a_m)\}_{m=1}^{M}$, where $s_m$ and $e_m$ represent the start and end time for the action class $a_m$, respectively.
    Different from conventional TAL that the model can access all frames in $V$, On-TAL allows the model to access $\{x_1, ..., x_t\}$ to produce the action class at time step $t$.
    Following prior works \cite{kang2021cag, wang2021oadtr}, we feed consecutive $l$ frames into the pretrained 3D CNNs (\eg TSN~\cite{}) to extract a $D_{in}$-dimensional visual feature $f_t \in \mathbb{R}^{D_{in}}$.
    On-TAL model takes $\{f_1, ...,f_t\}$ to produce the action $a_m$ at each time step $t$.
    }

\subsection{Motivation}
    \ji{To learn and apply the model in the online setting, we take some conditions into account to design the model:
    1) The context of the past information should be modeled as in \cite{kang2021cag};
    2) multiple action instances in the same time step should be detected;
    3) action instances should be detected in real-time for real-world applications.
    While \cite{kang2021cag} validated the effectiveness of context information, they failed to satisfy the other two conditions.}
    
    \ji{Building upon these motivations, we propose a simple yet effective framework, SimOn, for On-TAL using Transformer~\cite{vaswani2017attention}.
    Contrary to the prior work~\cite{kang2021cag} that uses the serial of outputs from the model as contexts, we leverage two context information as a full context, the past visual context which is a set of past visual features and their corresponding action probabilities, and an additional context embedding generated from a lightweight network.
    The past visual context can be used to learn the long-term context and the context embedding reflect the short-term context that the relation between the current frame and the consecutive past frame.
    Instead of predicting the actionness score for each frame, our model directly produces the action class probability, enabling multiple action predictions.
    As our model consists of lightweight architecture and takes only $k$ past contexts, the proposed method is carried out in real-time.}

\begin{figure}[t]
    \begin{center}
       \includegraphics[width=1\linewidth]{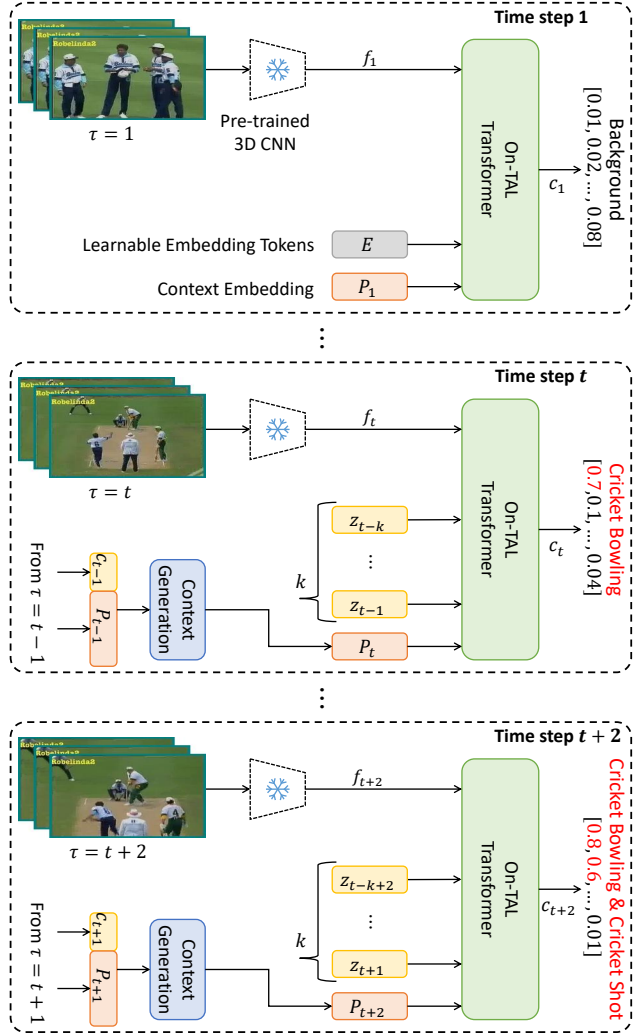}
    \end{center}
       \caption{The overall architecture of SimOn for On-TAL.}
    \label{fig:2}
    \end{figure}

		

\section{Proposed Method}
\subsection{SimOn}\label{subsection:simon}
    \ji{We describe SimOn in detail in the following sections.
    The primary objective of On-TAL is to correctly predict the action class probabilities to retrieve action instances $\psi$.
    To this end, we adopt a simple Transnformer~\cite{vaswani2017attention} that takes the current visual feature as the query and the past information as the keys and values.
    \paragraph{Initialization.}
    Since past information is not provided to the model at time step 1, we employ learnable embeddings as the past contexts to produce the reliable action probability for the first frame.
    Specifically, a learnable embedding $E\in\mathbb{R}^{D}$ and the first context embedding $P\in\mathbb{R}^{D}$ are randomly initialized.
    The model takes the stack of $E$ and $P_1$ as the keys and values and $f_1$ as the query as described below:
    \begin{gather}
        K = V = [E+PE, P_1+PE], \\
        Q = f_1 W_f + PE,
    \end{gather}
    where $PE$ denotes the positional encoding~\cite{vaswani2017attention} and $W_f \in \mathbb{R}^{D_{in} \times D}$ is a linear projection parameter.
    After predicting the action probability $c_1$ for the first frame, we concatenate $P_1$ and $c_1$ and project them into the $D$-dimensional feature space using two MLP followed by the ReLU activation and $\ell_2$-norm to make the context embedding for the second frame:
    \begin{equation}
        P_2 = \ell_2(\text{ReLU}(\text{Concat}(P_1, c_1)W_1)W_2),
    \end{equation}
    where $W_1$ and $W_2$ are the parameters in MLP.
    In addition, $Q$ and $c_1$ are concatenated, projected into the $D$-dimensional feature space, and utilized as the past visual context in the next time step, such that the keys and values in time step 2 are given by
    \begin{equation}
        K = V = [E+PE, z_1, P_2+PE],
    \end{equation}
    where $z_1 = \text{Concat}(Q;c_1W_c)W_o$ with learnable projection parameters $W_c$ and $W_o$.
    The learnable embedding $E$ remains as the component of the keys and values until $k$ past visual contexts are given to the model and popped out after $t = k$.
    \paragraph{At time step $t > k$.}
    The model takes $k$ past visual contexts and the context embedding at each time step.
    The keys, values, and query at time step $t>k$ are denoted as
    \begin{gather}
        K = V = [z_{t-k}, z_{t-k+1}, ..., z_{t-1}, P_t], \\
        Q = f_t W_f + PE,
    \end{gather}
    }
    \ky{where $z_t = \text{Concat}(Q;c_tW_c)W_o$.}

    \begin{figure}[t]
    \begin{center}
       \includegraphics[width=1\linewidth]{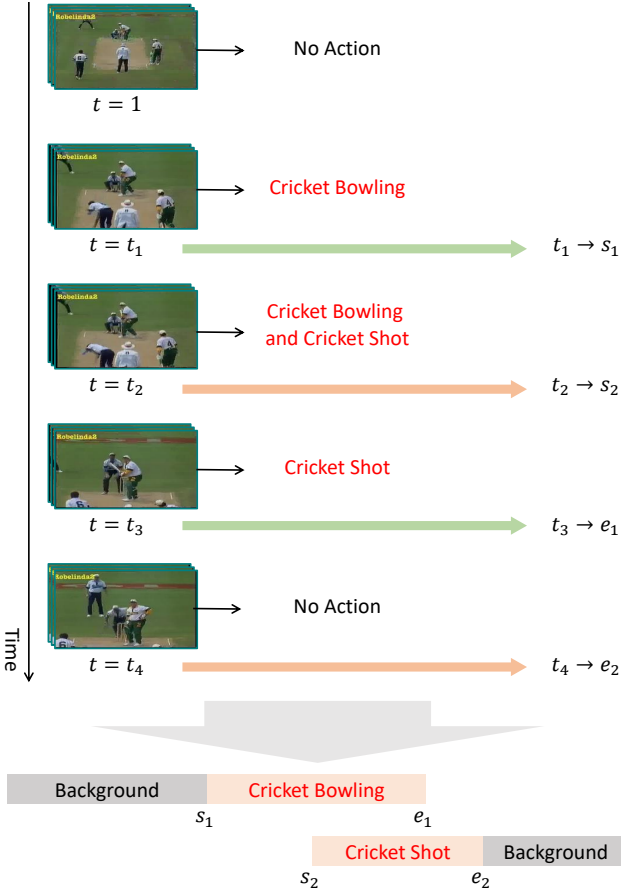}
    \end{center}
       \caption{Example of the multiple action instance prediction.}
    \label{fig:3}
    \end{figure}

    \ji{We adopt Multi-Head Attention (MHA) in Transformer that learns to augment semantic information from past contexts.
    Formally, the MHA is defined as:
    \begin{gather}
            \operatorname{MHA}(Q, K, V) = \operatorname{Concat}\left(\operatorname{head}_{1}, \ldots, \text{head}_{\mathrm{H}}\right) W^{O},  \\
            \text{head}_{\mathrm{h}} = \operatorname{Attention}\left(Q_{h} W_{h}^{Q}, K_{h} W_{h}^{K}, V_{h} W_{j}^{V}\right),  \\
            \operatorname{Attention}(Q_{h}^{\prime}, K_{h}^{\prime}, V_{h}^{\prime}) = \operatorname{softmax}\left(\frac{Q_{h}^{\prime} K_{h}^{\prime T}}{\sqrt{D_{h}}}\right) V_{h}^{\prime},
    \end{gather}
    where $H$ denotes the number of heads in the MHA, $K_{h}$, $V_{h}$ $\in \mathbb{R}^{(k+1)\times \frac{D}{h}}$, and $Q_{h}$ $\in \mathbb{R}^{\frac{D}{h}}$ are h-ways splitted along feature dimension.
    $W_{h}^{K}$, $W_{h}^{V}$, and $W_{h}^{Q}$ $\in \mathbb{R}^{\frac{D}{h}\times \frac{D}{h}}$ are learnable linear projection parameters.
    The scaling factor $D_h = \frac{D}{H}$ is used for the stable training.
    Subsequently, the output of the MHA is fed into the Feed Forward Network with the residual connection \cite{he2016deep}, layer norm \cite{ba2016layer}, and the ReLU activation \cite{agarap2018deep}:
    \begin{gather}
        Q' = \operatorname{LN}\left(\operatorname{MHA}\left(Q, K, V\right) + Q \right),\\
        \hat{Q} = \operatorname{LN}\left(\operatorname{FFN}\left(Q'\right) + Q'\right),\\
        \operatorname{FFN}(Q')= \operatorname{ReLU}(Q' W_{3}) W_{4},
    \end{gather}
    where LN denotes the layer norm and $W_3, W_4\in \mathbb{R}^{D\times D}$ are linear projection parameters.
    }

    \ji{Finally, the action probability for the $t$-th frame is derived from a simple linear projection followed by a sigmoid activation function:
    \begin{equation}
        c_t = \operatorname{Sigmoid}(\hat{Q}W_C),
    \end{equation}
    where $W_C\in\mathbb{R}^{D\times C}$ projects the output of the Transformer to the action label space $\mathbb{R}^{C}$.
    The overall procedure of SimOn is depicted in \figref{fig:2}.
    }

    \subsection{Action Instance Prediction}
    \ji{
    The prior work~\cite{kang2021cag} decided the time interval $(s_m, e_m)$ first and classified the action class $a_m$ corresponding to the time interval.
    Since they only determined whether the frame contains action at each time step, they can not perform multiple action instance prediction.
    Contrary to this, our SimOn directly predicts action probability $c_t$ and determines action instances based on $c_t$ by applying thresholding.
    For example, assume the video that have overlapped action instances $\psi = \{(s_1, e_1, a_1), (s_2, e_2, a_2)\}$, where $a_1$ and $a_2$ represent the different action categories and $s_1 < s_2 < e_1$, as shown in \figref{fig:3}.
    If one action is detected at time $t_1$, we set $t_1$ as $s_1$ and the detected action to $a_1$.
    When the action $a_1$ and the other action $a_2$ are simultaneously detected at time $t_2$, we set $t_2$ as $s_2$.
    Finally, the ending times $e_1$ and $e_2$ can be set to the time indices when $a_1$ and $a_2$ are no longer detected.
    }
\begin{table*}[]
    \centering
    \begin{tabular}{cccccccc}
    \hline
    \multirow{2}{*}{Setting}  & \multirow{2}{*}{Method}                                    & \multicolumn{6}{c}{tIoU}                                                                                                                                                                                                                                           \\ \cline{3-8} 
                              &                                                            & 0.3                            & 0.4                            & 0.5                            & 0.6                            & \multicolumn{1}{c}{0.7}                            & \multicolumn{1}{c}{Average}                                              \\ \hline
    \multirow{21}{*}{Offline} & S-CNN \cite{shou2016temporal}             & 36.3                           & 28.7                           & 19.0                           & 10.3                           & \multicolumn{1}{c}{5.3}                            & 19.9                                                                      \\
                              & SMS \cite{yuan2017temporal}               & 36.5                           & 27.8                           & 17.8                           & -                              & \multicolumn{1}{c}{-}                              & -                                                                         \\
                              & SSAD \cite{lin2017single}                 & 43.0                           & 35.0                           & 24.6                           & -                              & \multicolumn{1}{c}{-}                              & -                                                                         \\
                              & CDC \cite{shou2017cdc}                    & 40.1                           & 29.4                           & 23.3                           & 13.1                           & \multicolumn{1}{c}{7.9}                            & 22.8                                                                      \\
                              & SST \cite{buch2017sst}                    & 41.2                           & 31.5                           & 20.0                           & 10.9                           & \multicolumn{1}{c}{4.7}                            & 21.7                                                                      \\
                              & SSN \cite{zhao2017temporal}               & 51.9                           & 41.0                           & 29.8                           & -                              & \multicolumn{1}{c}{-}                              & -                                                                         \\
                              & BSN \cite{lin2018bsn}                     & 53.5                           & 45.0                           & 36.9                           & 28.4                           & \multicolumn{1}{c}{20.0}                           & 36.8                                                                      \\
                              & TAL-Net \cite{chao2018rethinking}         & 53.2                           & 48.5                           & 42.8                           & 33.8                           & \multicolumn{1}{c}{20.8}                           & 39.8                                                                      \\
                              & SS-TAD \cite{buch2019end}                 & 45.7                           & -                              & 29.2                           & -                              & \multicolumn{1}{c}{9.6}                            & -                                                                         \\
                              & GTAN \cite{long2019gaussian}              & 57.8                           & 47.2                           & 38.8                           & -                              & \multicolumn{1}{c}{-}                              & -                                                                         \\
                              & G-TAD \cite{xu2020g}                      & 54.5                           & 47.6                           & 40.2                           & 30.8                           & \multicolumn{1}{c}{23.4}                           & 39.3                                                                      \\
                              & G-TAD+P-GCN \cite{xu2020g}                & 66.4                           & 60.4                           & 51.6                           & 37.6                           & \multicolumn{1}{c}{22.9}                           & 47.8                                                                      \\
                              & A2Net \cite{yang2020revisiting}           & 58.6                           & 54.1                           & 45.5                           & 32.5                           & \multicolumn{1}{c}{17.2}                           & 41.6                                                                      \\
                              & RTD-Net \cite{tan2021relaxed}             & 68.3                           & 62.3                           & 51.9                           & 38.8                           & \multicolumn{1}{c}{23.7}                           & 49.0                                                                      \\
                              & VSGN \cite{zhao2021video}                 & 66.7                           & 60.4                           & 52.4                           & 41.0                           & \multicolumn{1}{c}{30.4}                           & 50.2                                                                      \\
                              & RTD-Net \cite{tan2021relaxed}             & 68.3                           & 62.3                           & 51.9                           & 38.8                           & \multicolumn{1}{c}{23.7}                           & 49.0                                                                      \\
                              & ContextLoc \cite{zhu2021enriching}        & 68.3                           & 63.8                           & 54.3                           & 41.8                           & \multicolumn{1}{c}{26.2}                           & 50.9                                                                      \\
                              & End-to-End learning \cite{yeung2016end}   & 36.0                           & 26.4                           & 17.1                           & -                              & \multicolumn{1}{c}{-}                              & -                                                                         \\
                              & AFSD \cite{lin2021learning}               & 67.3                           & 62.4                           & 55.5                           & 43.7                           & \multicolumn{1}{c}{31.1}                           & 52.0                                                                      \\
                              & TadTr \cite{liu2021end}                   & 62.4                           & 57.4                           & 49.2                           & 37.8                           & \multicolumn{1}{c}{26.3}                           & 46.6                                                                      \\
                              & ActionFormer \cite{zhang2022actionformer} & \textbf{75.5} & \textbf{72.5} & \textbf{65.6} & \textbf{56.6} & \multicolumn{1}{c}{\textbf{42.7}} & \textbf{62.6}                                            \\ \hline
    \multirow{3}{*}{Online}   & CAG-QIL \cite{kang2021cag}                & 44.7                           & 37.6                           & 29.8                           & 21.9                           & \multicolumn{1}{c}{14.5}                           & 29.7                                                                      \\ \cline{2-8} 
                              & SimOn (Our)                                                & 54.3                           & 45.0                           & 35.0                           & 23.3                           & \multicolumn{1}{c}{14.6}                           & 34.4 (\textcolor{red}{+4.7})                           \\
                              & SimOn (Our) *                                              & \textbf{57.0} & \textbf{47.5} & \textbf{37.3} & \textbf{26.6} & \multicolumn{1}{c}{\textbf{16.0}} & \textbf{36.9} (\textcolor{red}{+7.2}) \\ \hline
    \end{tabular}
\caption{Results on the THUMOS14 dataset with different tIoU thresholds.  Average mAP in [0.3,0.4,..,0.7] is reported. 
The top results are marked in bold. Our method achieves the best results in the online fashion and is able to beat some early works in an offline manner, both one-stage, and two-stage. 
(*) indicates that the result is validated using the On-TAL ground truth from classification annotation.}
\label{table:sota}
\end{table*}

\subsection{Training Method for SimOn}
    \ji{In our framework, the whole parameters of the model are updated at every time step.
    Specifically, we reduce the problem of multiple action prediction to a series of binary classification tasks.
    Given $C$ actions, the action probability $c_t$ can be considered as a set of $c_t^i$ which is a probability for action $a_i$, where $1\leq i \leq C$.
    We employ Focal Loss (FL)~\cite{lin2017focal} to train SimOn by accumulating all binary losses from $C$ labels:
    \begin{equation}
    \begin{aligned}
        FL(c_t) = &\sum_{i=1}^{C}{-\alpha y_t^i (1-c_t^i)^\gamma \log(c_t^i)}\\
                &- (1-\alpha)(1-y_t^i)(c_t^i)^\gamma \log(1-c_t^i)
    \end{aligned}
    \end{equation}
    where $y_t^i$ is the groundtruth for class $i$, $\alpha$, and $gamma$ are the balancing parameters.
    }

\section{Experimental Results}
    \ji{In this section, we validate the effectiveness of the proposed method for On-TAL across a range of datasets, including the THUMOS14~\cite{jiang2014thumos} and ActivityNet1.3~\cite{caba2015activitynet} datasets.
    In addition, we verify the robustness of our model to the online video tasks by evaluating ODAS performance on the THUMOS14 dataset.}


\subsection{Experimental Setup}
\label{ex:setup}
\paragraph{Datasets.}
    \ji{We validate our method on two popular datasets: THUMOS14~\cite{jiang2014thumos} and ActivityNet1.3~\cite{caba2015activitynet}.
    The THUMOS14 dataset contains 200 validation videos and 213 testing videos with 20 action classes.
    Each video contains on average 15 action instances.
    We trained the model using validation videos and measured the performance on testing videos.
    The ActivityNet1.3 dataset provides 19,994 untrimmed videos which are divided into train, validation, and testing sets by a ratio of 2:1:2.
    They contain 200 different types of activities and each video has on average 1.5 action instances.}

\paragraph{Features Extraction.}
Following \cite{kang2021cag, wang2021oadtr}, we use 6 consecutive frames as input of two-stream network TSN \cite{wang2016temporal} pre-trained on the Kinetics~\cite{carreira2017quo} dataset to extract feature. On Activitynet1.3, each video feature is rescaled to length 100 by linear interpolation. When we evaluate our model for ODAS on the THUMOS14 dataset, Activitynet pre-trained features are utilized as in \cite{kang2021cag, shou2018online, gao2019startnet}.

\paragraph{Evaluation Metric.}
    \ji{For On-TAL, we use mean average precision (mAP) at different temporal intersections over union (tIoU) to evaluate the model, following the standard conventions.
    We report mAP across all action categories given tIoU thresholds and averaged mAP for all tIoU.}

    \ji{For ODAS, we adopt point-level average precision (p-AP), following previous works~\cite{gao2019startnet, kang2021cag, gao2019startnet}.
    }

We validate the model on the ODAS task using p-mAP at difference offset tolerance by averaging point-level average precision (p-AP) as previous works \cite{gao2019startnet, kang2021cag, gao2019startnet}.
Prior to measuring the p-AP, the predictions for each category are sorted in decreasing order based on predicted confidence scores. A prediction is considered as true positive if it belongs to the same class as the ground truth and the difference between its predicted temporal and ground-truth temporal is less than an offset tolerance. We use 10 offset tolerances start from 1 to 10.

\begin{figure*}[t]
	\begin{center}
		\includegraphics[width=1.\linewidth]{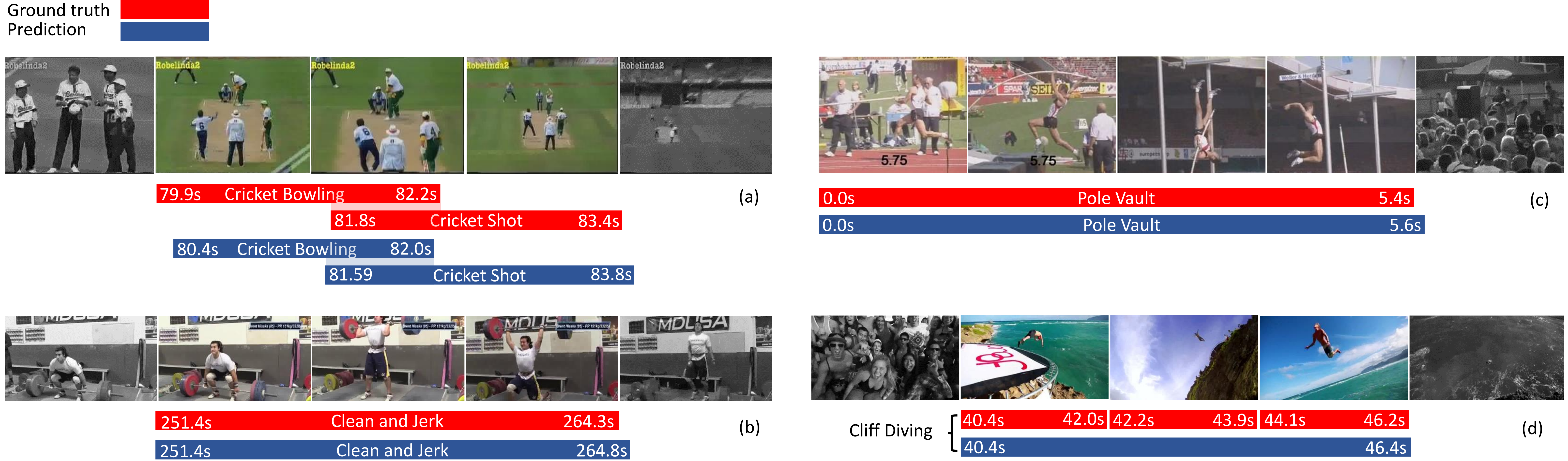}
	\end{center}
	\caption{Qualitative results of our proposed method from the THUMOS14~\cite{jiang2014thumos} dataset. Our model is able to (a) detect multiple actions at once, (b) localize accurately long-lasting action, and (c) prompt action at the beginning of the video. The inevitable action merging problem due to quantization error is shown in (d).}
	\label{fig:qualitative}
\end{figure*}
\begin{table}[]
    \centering
    \begin{tabular}{ccccc}
    \hline
    \multirow{2}{*}{Method} & \multicolumn{4}{c}{tIOU} \\
     \cline{2-5}
     & 0.5           & 0.75          & 0.9          &  Average      \\
    \hline
    CAG-BC \cite{kang2021cag} & 9.5          & 7.4          & 3.4          & 6.8          \\
    CAG-RL \cite{kang2021cag} & 21.4          & 11.3          & 2.2          & 11.6          \\
    CAG-QIL \cite{kang2021cag} & 30.5          & 18.5          & 4.1          & 17.7          \\
    \hline
    SimOn (Our) & 32.5          & 19.8          & 9.7          & 20.7 (\textcolor{red}{+3.0})          \\
    SimOn (Our)*      & \textbf{33.2} & \textbf{21.2} & \textbf{9.9} & \textbf{21.4} (\textcolor{red}{+3.7})\\
    \hline
    \end{tabular}
    \caption{Results on ActivityNet1.3. We show mAP at different tIoU thresholds. 
    Average mAP in [0.5,0.75,0.9] is reported. The best results are marked in bold. 
    Our method outperforms the latest method in an online manner by a large margin.
    (*) indicates that the result is validated using the On-TAL ground truth from classification annotation.}
    \label{table:annet}
    \end{table}
\begin{table*}[h]
	\centering
	\begin{tabular}{ccccccccc}
		\hline
		\multirow{2}{*}{\textbf{Method}}  & \multicolumn{6}{c}{tIOU} &\multirow{2}{*}{\#params}  &\multirow{2}{*}{inference time} \\
            \cline{2-7}
            & 0.3           & 0.4           & 0.5           & 0.6           & 0.7           & Average                    &  &     \\ \hline \hline
		CAG-QIL \cite{kang2021cag}       & 44.7          & 37.6          & 29.8           & 21.9           & 14.5          & 29.7        & -        & 78 ms                 \\ \hline
		OadTR \cite{wang2021oadtr}     & 47.1          & 39.1          & 31.7          & 22.7          & 13.5          & 30.8            & 75.8M    & 10.7 ms             \\ \hline
		SimOn w/o $P_{t}$     & 55.8          & 47.1          & 36.9          & 24.8          & 15.3          & 36.0 (\textcolor{red}{+5.2}) & 3.8M     & 4.9 ms               \\ 
		SimOn  & \textbf{57.0} & \textbf{47.5} & \textbf{37.3} & \textbf{26.6} & \textbf{16.0} & \textbf{36.9} (\textcolor{red}{+6.1})   & 3.94M     & 5.2 ms              \\ \hline
	\end{tabular}
	\caption{The effectiveness of our proposed methods on the model performance. 
	We report different results at different tIoU thresholds in [0.3, 0.4,..., 0.7] and average mAP on THUMOS14.
	It is worth noting that OadTR cannot detect action at the beginning of the video. And existing methods cannot detect more than one action at the same time.
	The last two columns show the number of parameters (\#params) of the model and inference speed (ms) respectively. 
	The inference time is reported with batch size 1 (stack 6 consecutive frames) on a single GeForce GTX 1080Ti.}
	\label{table:hif}
\end{table*}

\paragraph{Implementation details.}
    \ji{For On-TAL Transformer, we used 4 Transformer blocks and each block consists of MHA with 8 heads and FFN.
    We set the dropout ratio to 0.1 while training.
    We train the model with a batch size of 256 for 16 epochs using AdamW optimizer~\cite{kingma2014adam} with a step learning rate scheduler, step size of 3.
    The base learning rate for the On-TAL Transformer and the context embedding generation are set to 1e-4 and 1e-5, respectively.}

\paragraph{Observation on groundtruth.}
\ky{When evaluating the model on On-TAL, we notice that the model inherently can not achieve 100 mAP. There are two reasons for the performance degrading. The first one is the quantization error. \ky{Since we take newly arrived $l$ consecutive frames for each visual feature, as $l$ increases,} the more quantization error we encounter. The second reason is action merging. It happens when actions belong to the same category, and the end time of one action is the start time of another action. In this case, the action grouping process will merge those actions into a single one. We illustrates the action merging in figure \ref{fig:qualitative}(d).}
\ky{Hence, to be able to measure performance of the model without aforementioned inherent performance degradings, we create a new On-TAL ground turth from classification annotation by applying quantization and merging in advance.}
The result with the new On-TAL ground truth is marked with * in the experimental section.

\subsection{Main Results}
\label{ex:result_ontal}

\subsubsection{Experimental Results for On-TAL} 
\paragraph{Results on THUMOS14}
    \ji{We compare the performance evaluated on the THUMOS14~\cite{jiang2014thumos} with comprehensive state-of-the-art online approach~\cite{kang2021cag} and offline approaches~\cite{} in \tabref{table:sota}, including online and offline methods.
    The results show our model outperforms the prior work~\cite{kang2021cag} for the online fashion at every single tIoU threshold, demonstrating the effectiveness of the proposed method.
    Especially, our model achieves 57.0\% mAP at tIoU 0.3, surpassing CAG-QIL~\cite{kang2021cag} by a large margin (+12.3mAP).
    The comparisons between our method and offline approaches show that the proposed method achieves competitive performance, further reducing the gap between online and offline learning.}

\paragraph{Result on ActivityNet1.3.} 
    \ji{We evaluate the performance on the ActivityNet1.3~\cite{caba2015activitynet} dataset.
    As shown in \tabref{table:annet}, our method outperforms \cite{kang2021cag} at every tIoU thresholds.
    Especially, our model improves 5.8mAP at tIoU=0.9.
    On average, we achieve 21.4mAP, which is 3.7mAP higher than the previous approach, demonstrating the robustness of our method across datasets.
    }

\paragraph{Qualitative results.}
    \ji{We provide qualitative results according to the various challenging conditions, including overlapped multiple action instances, long-lasting actions, and promptly started actions.
    Our model produces confident predictions for overlapped multiple action instances as shown in \figref{fig:qualitative}-(a), showing different actions in the same time step can be detected separately.
    In addition, our model can accomplish to predict relatively long-lasting actions, proving the effectiveness of the context modeling as shown in \figref{fig:qualitative}-(b).
    The initial learnable parameter enables the model to detect prompt actions at the beginning of the video as shown in \figref{fig:qualitative}-(c).}

\paragraph{Efficiency analysis.}
    \ji{To demonstrate the computation and memory efficiency, we compare the number of model parameters and inference time with state-of-the-art On-TAL method~\cite{kang2021cag}.
    In addition, we provide the comparison between our model and state-of-the-art OAD model~\cite{wang2021oadtr} to verify the applicability of the OAD model to On-TAL.
    As shown in \tabref{table:hif}, we report the On-TAL performance evaluated on the THUMOS14 dataset, the number of model parameters, and inference time.
    The comparisons between our method and other approaches show that the proposed method not only provides outstanding performance but is also efficient in terms of computation and memory costs.
    While the OadTR~\cite{wang2021oadtr} comprises 75.8M parameters, our model contains only 3.94M of parameters, requiring about 20x less memory than OadTR.
    For the inference time, our model takes 5.2ms for stacked 6 consecutive frames, which is 13.2x and 2.1x faster than \cite{kang2021cag} and \cite{wang2021oadtr}, respectively.
    Note that we measure the inference time using a single Geforce 1080Ti GPU.}

\subsubsection{Experimental Results for ODAS}
\label{ex:odas}
    \ji{To verify the applicability of our method to the online video tasks, we evaluate our model on the THUMOS14~\cite{jiang2014thumos} dataset for the ODAS task.
    We provide the comparisons between our method and state-of-the-art ODAS approaches in \tabref{table:odas}.
    The results show that our SimOn remarkably outperforms state-of-the-art models on every offset tolerance with large margins.
    While we do not train the model for ODAS, our model achieves 15.2p-mAP performance improvement on average, validating the effectiveness of the proposed framework on the online video tasks.
    }


\begin{table*}[t]
	\centering
	\resizebox{\textwidth}{!}{ %
		\begin{tabular}{ccccccccccc}
            \hline
  
			\multirow{2}{*}{Method} & \multicolumn{10}{c}{Offsets} \\
            \cline{2-11}
            & \textbf{1} & \textbf{2} & \textbf{3} & \textbf{4} & \textbf{5} & \textbf{6} & \textbf{7} & \textbf{8} & \textbf{9} & \textbf{10} \\
			\hline
			Shou et al. \cite{shou2018online}  & 3.1        & 4.3        & 4.7        & 5.4        & 5.8        & 6.1        & 6.5        & 7.2        & 7.6        & 8.2         \\
			ClsNet-only \cite{gao2019startnet} & 13.9       & 21.6       & 25.8       & 28.9       & 31.1       & 32.5       & 33.5       & 34.3       & 34.8       & 35.2        \\
			StartNet-CE \cite{gao2019startnet} & 17.4       & 25.4       & 29.8       & 33.0       & 34.6       & 36.3       & 37.2       & 37.7       & 38.6       & 38.8        \\
			StartNet-PG \cite{gao2019startnet} & 19.5       & 27.2       & 30.8       & 33.9       & 36.5       & 37.5       & 38.3       & 38.8       & 39.5       & 39.8        \\
			CAG-QIL \cite{kang2021cag}         & 20.3       & 31.2       & 37.2       & 41.4       & 44.2       & 46.0         & 47.3       & 48.1       & 48.9       & 49.8      \\
			\hline
			SimOn (Our) 					           & 27.3       & 41.2       & 48.6       & 54.0       & 57.5       & 59.3         & 60.9       & 62.7       & 63.1       & 63.8 \\
			SimOn(Our)*                             & \textbf{28.0}{\tiny (\textcolor{red}{+7.7})} & \textbf{41.7}{\tiny (\textcolor{red}{+10.5})}  & \textbf{49.1}{\tiny (\textcolor{red}{+11.9})} & \textbf{54.8}{\tiny (\textcolor{red}{+13.4})} & \textbf{58.4}{\tiny (\textcolor{red}{+14.2})} & \textbf{60.4}{\tiny (\textcolor{red}{+14.4})} & \textbf{61.8}{\tiny (\textcolor{red}{+14.5})} & \textbf{63.7}{\tiny (\textcolor{red}{+15.6})} & \textbf{64.3}{\tiny (\textcolor{red}{+15.4})} & \textbf{65.0}{\tiny (\textcolor{red}{+15.2})}\\
			\hline
			
		\end{tabular}
	}
	\caption{Experiments results on THUMOS14 on online detection of action start (ODAS). It is worth noting that our model is not intentionally trained for ODAS, but we accomplish new SOTA result on ODAS task. 
	(*) indicates that the result is validated using the On-TAL ground truth from classification annotation.}
	\label{table:odas}
\end{table*}

\subsection{Ablation Study}
\label{ex:ablation}
    \ji{We conduct various ablation studies to verify the effects of components of SimOn, including the number of past contexts $k$ and the value of threshold to determine the action class.
    Note that all experiments in this section are conducted on the validation set of the THUMOS14 dataset and the performance is evaluated by using the On-TAL ground truth from classification annotation.}

\subsubsection{Different Values of $k$}
\label{subsubsection:ws}
\begin{figure}
    \centering
    \begin{subfigure}[b]{0.48\linewidth}        
        \centering
        \includegraphics[width=\linewidth]{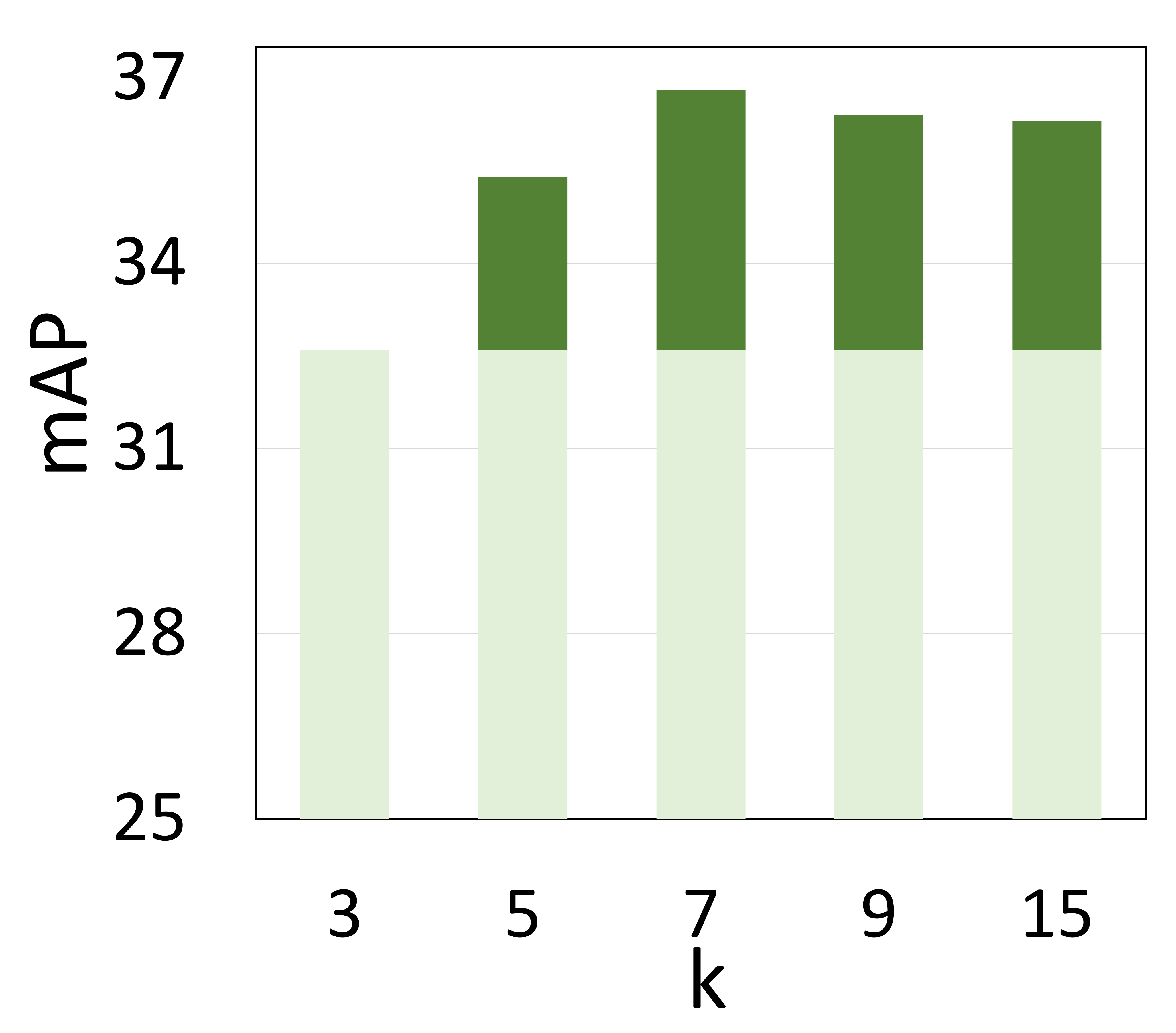}
        \caption{$k$ vs. mAP}
        \label{fig:k}
    \end{subfigure}
    \begin{subfigure}[b]{0.48\linewidth}        
        \centering
        \includegraphics[width=\linewidth]{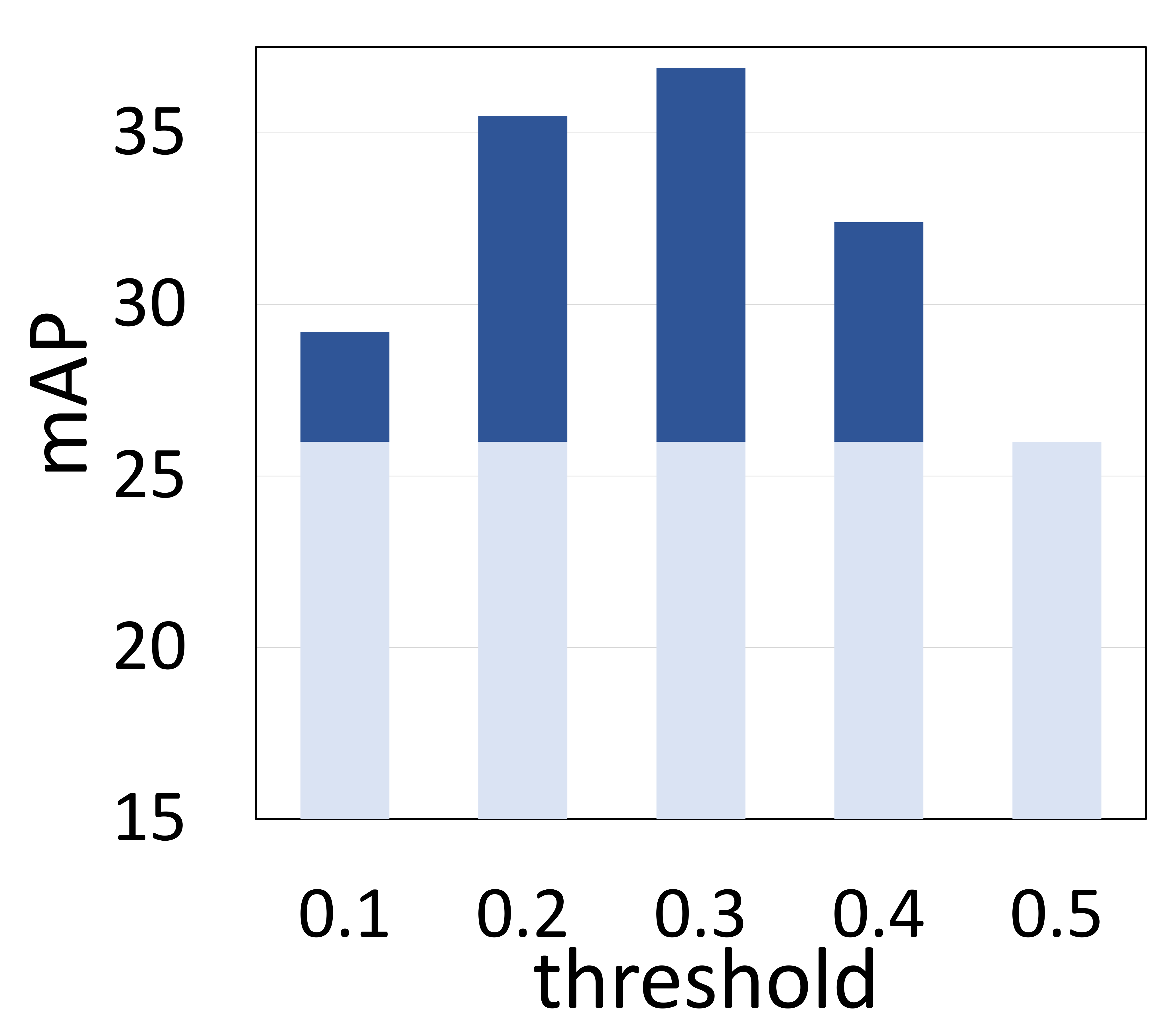}
        \caption{$thresholds$ vs. mAP}
        \label{fig:thr}
    \end{subfigure}
    \caption{Ablation study about the number of the latest feature and classification probabilities $k$ (a) and $thresholds$ (b). The average mAP on THUMOS14 test set is reported.}
    \label{fig:ablation}
\end{figure}

    \ji{We investigate how the quantity of past context information (\ie the number of keys and values) affects the capacity of the model.
    The performance according to various $k$ is reported in figure \ref{fig:ablation}-(a). 
    Our model achieves the highest performance with the value of $k=7$ and the lowest performance appeared with the value of $k=3$.
    Since the model predicts the action probability by interacting with the past contexts, the high performance can not be obtained with a small value of $k$.
    Meanwhile, a large number of $k$ induce a higher computational cost, making the training procedure inefficient.
    In addition, distant past context from the current frame may harm the capacity of the model.
    In practice, we use the value of $k=7$ in the main experiments.}



\subsubsection{Different Values of thresholds}
\label{subsubsection:thr}

Moving further, we investigate the impact of the $threshold$ value on performance as it is a crucial hyper-parameter for the decision of action start/end. 
We vary this parameter and illustrate the results in figure \ref{fig:ablation}-(b). Our method gives the best result at $threshold=0.3$, and gradually degrades around the optimal threshold. The model tends to produce quite low scores due to the utilization of focal loss \cite{lin2017focal}, in which the loss down-weights easy samples and focuses on the hard samples while training. Hence, the scores for a class may not be high.

\section{Conclusion and Future work}
    In this paper, we propose a simple On-TAL framework, termed SimOn, that effectively learns past context information to enable the model to predict action instances of a video in an end-to-end manner.
    In our framework, we formulate past context information as long-term and short-term contexts to make the model robust to the various challenging conditions. 
    The experimental analysis demonstrated the superiority of the proposed method, achieving a new state-of-the-art performance for On-TAL and ODAS tasks.
    In addition, we validate the effectiveness of the components of SimOn through extensive ablation studies.
    We hope that our simple approach will provide a solid baseline for future research on online video tasks.

    While SimOn has attained competitive performance compared with the offline approaches that can access to future frames and apply post-processing to the output of the model, there is still a performance gap to be reduced.
    As the next step, we will investigate a novel context that can be incorporated into On-TAL framework.



{\small
	\bibliographystyle{ieee_fullname}
	\bibliography{egbib}
}

\end{document}